\documentclass[runningheads]{llncs}
\usepackage{cite}
\usepackage{graphicx}
\usepackage{amsmath}
\usepackage{verbatim}
\usepackage{multirow}
\usepackage{amssymb}
\usepackage{caption}
\usepackage{floatrow}
\usepackage{subfigure}
\usepackage{multirow}
\usepackage{comment}
\usepackage{siunitx}
\usepackage{booktabs}
\usepackage{color}
\usepackage{hyperref}
\hypersetup{
    colorlinks=true,
    linkcolor=blue,
    filecolor=magenta,
    urlcolor=red,
}
\usepackage{bm}
\usepackage[utf8]{inputenc}

\newfloatcommand{capbtabbox}{table}[][\FBwidth]
\begin{document}
\title{Smart Home Device Detection Algorithm Based on FSA-YOLOv5}
\author{Jiafeng Zhang\textsuperscript{1}, Xuejing Pu\textsuperscript{2}}

\authorrunning{Jiafeng Zhang}

\institute{
\textsuperscript{1}Tianjin Normal University\\
\textsuperscript{2}Southeast University\\ }
\maketitle
\begin{abstract}
Smart home device detection is a critical aspect of human-computer interaction. However, detecting targets in indoor environments can be challenging due to interference from ambient light and background noise. In this paper, we present a new model called FSA-YOLOv5, which addresses the limitations of traditional convolutional neural networks by introducing the Transformer to learn long-range dependencies. Additionally, we propose a new attention module, the full-separation attention module, which integrates spatial and channel dimensional information to learn contextual information. To improve tiny device detection, we include a prediction head for the indoor smart home device detection task. We also release the Southeast University Indoor Smart Speaker Dataset (SUSSD) to supplement existing data samples. Through a series of experiments on SUSSD, we demonstrate that our method outperforms other methods, highlighting the effectiveness of FSA-YOLOv5.

\end{abstract}

\section{Introduction}
As smart home devices continue to gain popularity, people are increasingly demanding higher performance and better user experience To meet these expectations, the application of human-computer interaction technology in the smart home field has become essential. This technology provides accurate, natural, and convenient interaction, improving device usability and ease of use. Additionally, it enables smart home devices to better understand users' needs and intentions, offering intelligent, efficient, and personalized services that ultimately enhance their quality of life.

To enhance the precision and convenience of human-computer interaction, computer vision technology is essential for visual detection. On the user's end, detecting their behavior and requirements is crucial for smart home devices to respond accurately and provide customized services. Additionally, smart home devices require visual detection to acquire precise information about the status and surroundings of the interacting devices. This enables the devices to operate adaptively and optimize the user experience automatically.

Convolutional neural network (CNN)~\cite{deep,deep2} has excellent characterization ability, making CNN-based methods the mainstream in the field of detection. CNN-based detection methods are categorized into two types: one-stage methods and two-stage methods~\cite{rcnn}. In the two-stage method, regions of interest (ROIs) are generated using selective search, and CNN is used to extract features for localization and classification of each ROI, as seen in R-CNN~\cite{rcnn}. However, achieving real-time performance with two-stage detection methods can be challenging due to the model's high complexity. In contrast, single-stage methods like the YOLO series~\cite{yolov3,YOLO,tph,yolov4} have excellent real-time performance, but detection accuracy needs improvement.

In this paper, we propose a new algorithm called FSA-YOLOv5 for smart home device detection tasks. The full separation attention module (FSA) is introduced in the FSA-YOLOv5 of Neck to integrate spatial and channel information and obtain comprehensive contextual information. Furthermore, we release the Southeast University Indoor Smart Speaker Dataset (SUSSD) to complement indoor smart home device datasets. Finally, we conduct a series of experiments on SUSSD, and the experimental results validate the effectiveness of our proposed FSA-YOLOv5 algorithm.

\section{Method}

In this section, we present our proposed FSA-YOLOv5 algorithm, which utilizes fully separated attention modules to learn key semantic information for improving the representational capability of the model. We first introduce the overall framework of FSA-YOLOv5, and then provide detailed descriptions of the backbone, neck, and head networks, including their respective functions and roles.

\begin{figure*}[t!]
    \centering
    \includegraphics[width=\textwidth]{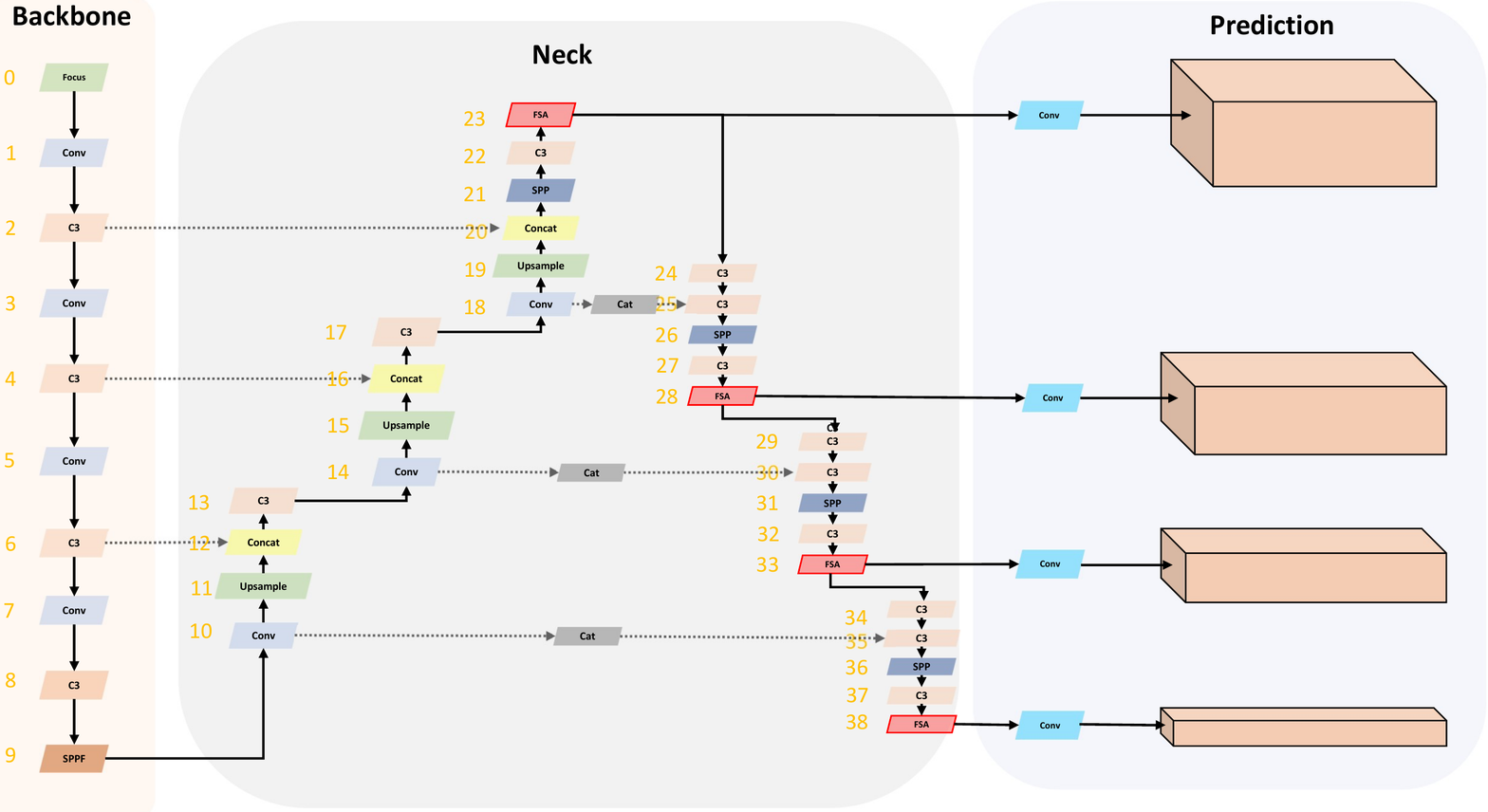}
    \caption{The overall framework of the proposed FSA-YOLOv5.}
    \label{fig:framework}
\end{figure*}

Figure~\ref{fig:framework} illustrates the overall framework of FSA-YOLOv5, which includes backbone, neck, and head networks. The backbone network extracts information from the input image and converts the raw pixel information into representational features at different scales. The neck network further processes and combines the features yielded by the backbone network at different resolution and semantic levels to improve the accuracy and robustness of network. Finally, the head network utilizes the fused features to generate object location and category.

In FSA-YOLOv5, the CSPDarknet53~\cite{cspnet} network is employed as the backbone, but we modify some modules to satisfy the needs of smart home device detection and improve the model performance. The channel attention mechanism enhances the capacity of deep neural networks to capture inter-channel relationships, which results in better model performance. By assigning weights to each channel, the mechanism determines the contribution of each channel to the final output, allowing the model to select relevant features more effectively. However, the channel attention mechanism does not consider the relative location information as spatial information is integrated, leading to a lack of connections between different locations.

The spatial attention mechanism enhances the capacity of deep neural networks to capture inter-spatial relationships in input data, leading to better identification of the relevant parts. By assigning weights to each spatial location, the mechanism enables the model to focus on the crucial locations for the task. However, the spatial attention mechanism integrates channel information to determine the relationship between each location, which may result in the lack of interactions among channel information.


\begin{figure*}[t]
    \centering
    \includegraphics[width=\textwidth]{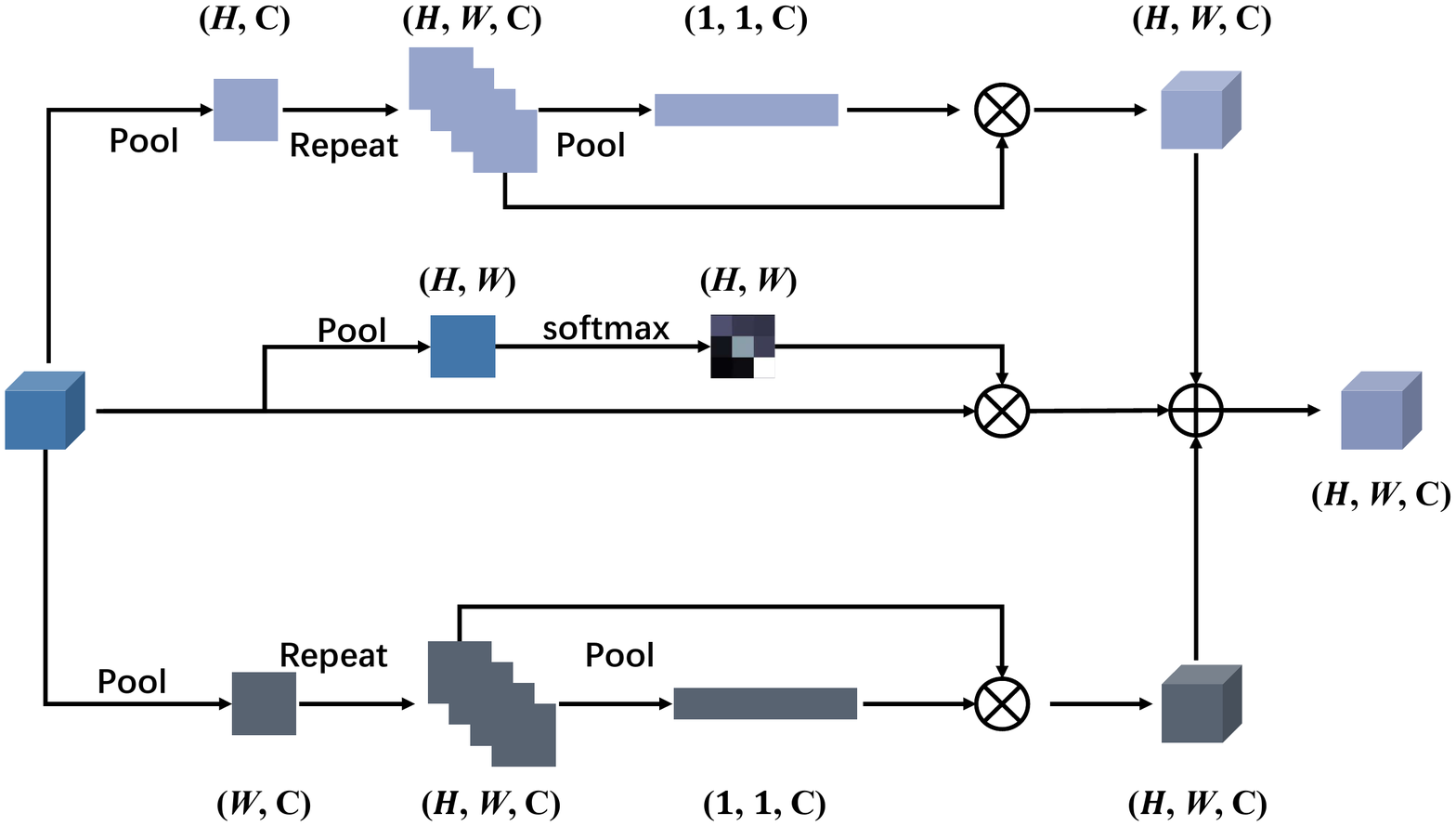}
    \caption{The structure of the fully separated attention module.}
    \label{fig:vis}
\end{figure*}

To address this limitation, we propose the fully separated attention module (FSA), which combines the channel attention mechanism and spatial attention mechanism. This module can capture inter-spatial relationships while preserving the interaction of channel information. Specifically, the channel attention mechanism determines the contribution of each channel by assigning weights, while the spatial attention mechanism assigns weights to each spatial location in the network. This combined approach enables better consideration of the relationship between different locations and channels in the input data, resulting in improved model performance and expressiveness.

The FSA module structure is illustrated in Figure~\ref{fig:vis}, which combines the spatial and channel dimension information of the feature map to obtain contextual information. The input features are split into three branches corresponding to the height, width, and number of channels of the feature map, denoted as H, W, and C, respectively. Each branch is globally pooled for the height, width, and channel dimensions, respectively, resulting in three feature maps. The branch feature maps of pooled height and width are duplicated multiple times to obtain feature maps of the same size as the input. The feature maps with pooled heights and widths are duplicated several times to match the size of the input feature map. Then, channel attention is applied to each replicated feature map to capture channel information. To complement spatial information lost in the first two branches, spatial attention is used to model relative position relationships for the pooled channel features. These feature maps are then merged to form the output of the hybrid attention module. The module calculates the importance of each pixel in the image using the hybrid attention mechanism and adjusts the feature maps to better capture global contextual information.

The FSA-YOLOv5 model utilizes FSA in neck to generate feature maps, which are then fed into the head network. The head network converts these feature maps into bounding boxes and applies non-maximal suppression (NMS) to filter out duplicate boxes, resulting in the final detection results.

To improve the accuracy of indoor home device detection, it is important to address the challenge of detecting small home furnishings and decorative items that are difficult to identify using traditional object detection algorithms. One way to overcome this challenge is by incorporating a specialized head network designed specifically for small object detection. This head network is capable of improving the detection of small objects by enhancing the representation information of shallow features. Despite the fact that this approach may increase computational effort, it is crucial to ensure the effectiveness of smart home detection. Therefore, a small-object detection head is being included in the prediction network in FSA-YOLOv5 to capture the feature information of these small objects.

\section{Experiments and Discussion}
\subsection{Dataset and Evaluation}

To comprehensively evaluate the performance of FSA-YOLOv5 for advanced object detection, we compared it to other object detection algorithms, including YOLOv3~\cite{yolov3}, Ghost-YOLOv5, YOLOv5, and TPH-YOLOv5~\cite{tph}. The results of this comparison are presented in Table 1.

Based on the results presented in Table 1, FSA-YOLOv5 outperforms all other advanced algorithms in all evaluation metrics. Specifically, FSA-YOLOv5 exhibits higher mAP@0.5, and mAP@0.5:0.95 compared to the second ranked algorithm, with improvements of 3.01\% and 3.37\%, respectively. It is worth noting that transform network-based algorithms, such as FSA-YOLOv5 and THP-YOLOv5, tend to outperform convolutional neural network-based algorithms due to their ability to more effectively model long-distance dependencies.

\begin{figure*}
    \centering
    \includegraphics[width=0.85\textwidth]{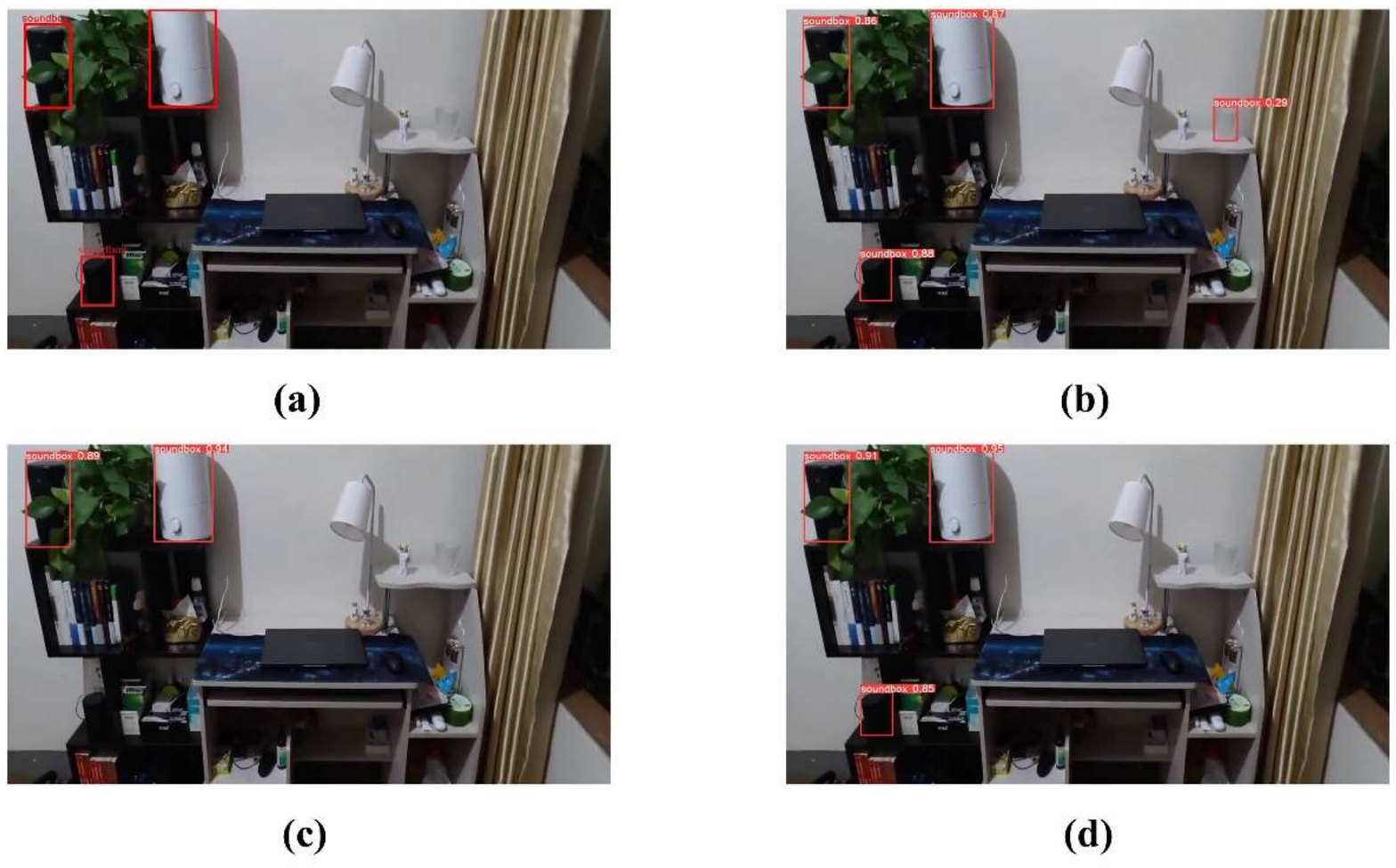}
    \caption{Visualization of the results of different detection algorithms. (a) Smart speaker sample. (b) TPH-YOLOv5. (c) YOLOv5. (d) FSA-YOLOv5.}
    \label{fig:abl:skip}
\end{figure*}

To perform qualitative analysis of the different algorithms, we visualize our detection results alongside those of other object detection algorithms in Figure~\ref{fig:abl:skip}. As shown in the figure, our method outperforms other advanced detection algorithms in detecting smart speakers indoors, highlighting the superiority of our approach.

%
%

\begin{table}
\caption{Comparison with other methods on SUSSD}
\label{Table2}
\centering
\begin{tabular}{l|l l l l}
\hline
Method       & Precision (\%) & Recall (\%) & map@0.5 (\%) & map@0.5:0.95 (\%)  \\
\hline
YOLOv3       & 90.95          & 75.34       & 80.62        & 59.75              \\
\hline
YOLOv5       & 94.88          & 77.67       & 81.90        & 65.99              \\
\hline
Ghost-YOLOv5 & 95.35          & 76.26       & 83.95        & 67.29              \\
\hline
TPH-YOLOv5   & 91.58          & 86.05       & 90.91        & 77.19              \\
\hline
FSA-YOLOv5    & 95.13          & 91.16       & 94.56       & 80.55              \\
\hline
\end{tabular}
\end{table}

\section{Conclusion}
In this paper, we have presented a novel method, FSA-YOLOv5, for detecting indoor smart home devices. Our method is comprised of backbone, neck, and head networks, and we introduce two key modifications to enhance its performance. Firstly, we integrate a transform network to capture long-distance dependencies and overcome the limitations of traditional convolutional neural networks in learning global information. Secondly, we propose a new attention module, the full separation attention module, in the neck network. This module effectively integrates spatial and channel dimension information of the feature map, enabling it to capture contextual information. Our experiments on this dataset demonstrate the superior performance of FSA-YOLOv5

%
%
%

\bibliographystyle{IEEEtran}
\bibliography{my}
%

\end{document}